\documentclass{article}
\usepackage{spconf,amsmath,graphicx}

\usepackage{enumitem}
\setlist{nosep, leftmargin=14pt}

\usepackage{mwe} 

\title{Bridging 3D Deep Learning and Curation for Analysis and High-Quality Segmentation in Practice}
\name{\parbox{\textwidth}{\centering
Simon Püttmann$^{1}$, Jonathan Jair Sánchez Contreras$^{1}$, Lennart Kowitz$^{1}$, Peter Lampen$^{1}$ \\ Saumya Gupta$^{2}$, Davide Panzeri$^{1}$, Nina Hagemann$^{3}$, Qiaojie Xiong$^{2}$, Dirk M. Hermann$^{3}$ \\ Chao Chen$^{2}$, Jianxu Chen$^{1\dagger}$ \thanks{$^{\dagger}$ Corresponding author, E-mail: jianxu.chen@isas.de}}
\address{
$^{1}$ Leibniz-Institut für Analytische Wissenschaften - ISAS - e.V., Dortmund, Germany\\
$^{2}$ Stony Brook University, New York, USA \\
$^{3}$ Department of Neurology, University Hospital Essen, University of Duisburg-Essen, Essen, Germany \\
}}

\begin{document}
%
\maketitle
\begin{abstract}
Accurate 3D microscopy image segmentation is critical for quantitative bioimage analysis but even state-of-the-art foundation models yield error-prone results. Therefore, manual curation is still widely used for either preparing high-quality training data or fixing errors before analysis. We present VessQC, an open-source tool for uncertainty-guided curation of large 3D microscopy segmentations. By integrating uncertainty maps, VessQC directs user attention to regions most likely containing biologically meaningful errors. In a preliminary user study
uncertainty-guided correction significantly improved error detection recall from $67\%$ to $94.0\%$ ($\text{p}\text{=}0.007$) without a significant increase in total curation time. VessQC thus enables efficient, human-in-the-loop refinement of volumetric segmentations and bridges a key gap in real-world applications between uncertainty estimation and practical human-computer interaction. The software is freely available at github.com/MMV-Lab/VessQC.

\end{abstract}
\begin{keywords}
3D Microscopy, Segmentation Curation, Uncertainty-Guided, Topology-Aware, napari-Plugin
\end{keywords}

\section{Introduction}
\label{sec:intro}
Accurate segmentation of 3D microscopy images is essential for the quantitative analysis of complex biological structures such as vascular networks \cite{chen20243d}. However, it remains one of the major bottlenecks in bioimage analysis \cite{wen20213deecelltracker, kousaka2025automated}. Despite remarkable advances in data-driven image analysis methods, 3D biomedical datasets often exhibit substantial variability in image quality, contrast, resolution, and tissue-specific morphology \cite{zhang2020generalizing, wittmann2025vesselfm}. These differences, together with domain gaps between hospitals, imaging protocols, artifacts, patient populations, etc., limit the generalization of supervised models and frequently lead to segmentation errors \cite{zhang2020generalizing, wittmann2025vesselfm}, particularly in thin, branching, or topology-sensitive structures such as vessels \cite{gupta2023topology}. Correcting these errors is critical for both biological interpretation and the creation of high-quality training datasets, yet it is also one of the most time-consuming steps in current analysis workflows in real-world applications.

\textbf{Automatic segmentation:} Recent developments in deep learning–based segmentation of biomedical images, including both 2D and 3D modalities, such as retinal fundus photography, confocal and light-sheet microscopy, have witnessed remarkable progress \cite{chen20243d}. Despite these advances, most existing methods focus exclusively on automatic segmentation, assuming access to sufficient ground-truth annotations and rarely address the subsequent quality control or curation stages that are essential for generating reliable results. Specific to vascular imaging, several open-source frameworks have lowered the entry barrier for automatic segmentation and analysis. Tools such as tUbeNet \cite{holroyd2023tube} and VesselMetrics \cite{mcgarry2024vessel} provide pipelines for vessel extraction, quantification, and feature analysis, while proprietary commercial software (e.g. Imaris Filament Tracer, Amira-Avizo XFibre) enables limited manual editing of segmented structures \cite{holroyd2023tube}. However, these solutions are designed primarily for inference or visualization and lack mechanisms for guided correction based on model confidence or structural integrity.

\textbf{Uncertainty estimation:} Recent research has increasingly focused on uncertainty estimation and segmentation quality control. Techniques such as Monte Carlo dropout, test-time augmentation, and model ensembles are used to quantify prediction confidence at the pixel or voxel level \cite{lin2022novel, mcgarry2024vessel}. Frameworks such as SegQC \cite{specktor2025segqc}, fuzzy uncertainty modeling \cite{lin2022novel}, and Uncertainty-Guided Annotation \cite{khalili2024uncertainty} have shown that uncertainty can effectively guide selective human review, improving dataset quality with minimal manual effort. But, how can biomedical researchers use these methods in practice? Most existing methods have been developed for 2D imaging modalities and typically stop at predicting uncertainty, without providing tools for interactive correction. 

\textbf{Topology-aware analysis:} Beyond voxel-level confidence, biological interpretability often depends on the topology of segmented structures: whether vessels remain connected, branches are properly formed, or loops are preserved. To address these limitations topology-aware loss functions and evaluation metrics have been introduced to better capture structural accuracy by identifying and quantifying errors in individual branches and connections rather than on a per-pixel basis \cite{gupta2023topology}. Yet, despite their conceptual advances, these approaches remain largely absent from practical tools that link structural feedback to visualization or correction. As a result, no existing framework unifies voxel-level uncertainty with topology-aware insights to support interactive, user-guided curation of 3D microscopy datasets.

\textbf{Data-centric AI:} 
In recent years, the field of bioimage analysis has increasingly embraced a data-centric perspective, which emphasizes that the quality of the data, rather than model complexity alone, determines the performance and reliability of downstream models \cite{cao2025rethinking}. From this viewpoint, improving datasets through targeted curation, correction, and enrichment is as crucial as developing new network architectures. Within this framework, one key stage (\textit{hunt-for-mistakes}) is to systematically identify and correct errors in existing predictions to iteratively refine datasets and improve model robustness \cite{cao2025rethinking}. In practice, however, this phase remains largely manual and inefficient, especially for large 3D microscopy volumes. While human-in-the-loop workflows, which combine automation with user feedback, are now common in 2D segmentation tasks, comparable solutions for 3D biomedical volumes remain underdeveloped \cite{zhou2023volumetric}. The combination of massive data sizes, complex 3D structures, and the absence of integrated guidance forces researchers to rely on exhaustive visual inspection, making it difficult to efficiently detect and prioritize critical segmentation errors.

In summary, prior work has established strong foundations for automatic segmentation, uncertainty estimation, topology-aware analysis, and dataset quality, but these components remain isolated. There is currently no open, interactive, uncertainty-guided environment that supports efficient post-segmentation correction in 3D microscopy, a capability that has been highlighted as a desirable next step for the vascular imaging community \cite{holroyd2023tube}. 

\section{Uncertainty-Guided Curation Tool (3D)}
\label{sec:tool}
\begin{figure}[htb]
  \centering
  \centerline{\includegraphics[width=8.5cm]{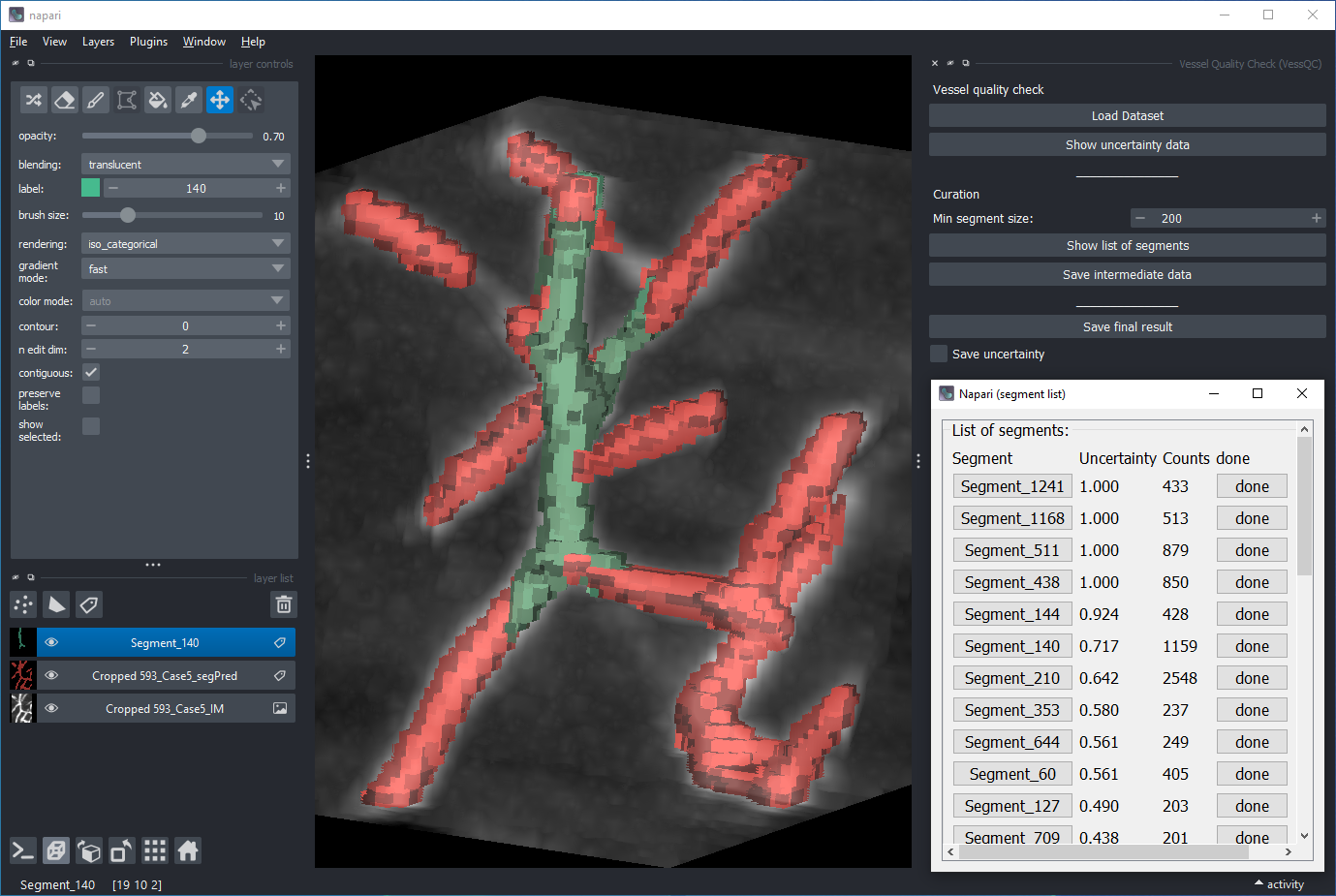}}
  \caption{VessQC Interactive Curation Interface (Napari). The interface displays a selected region extracted from the full 3D segmentation with the chosen branch visually highlighted. User tools for image modification (e.g., brightness, contrast) and view control (2D/3D toggle) are provided. The highlighted branch can be directly edited in 2D. Once edits are complete, changes are saved via the “Done” button, and the next branch for review can be immediately selected from the segmentation list on the right.}
  \label{fig:interface}
\end{figure}

In this work, we present VessQC, an open-source tool, for uncertainty-guided curation of large 3D microscopy image segmentations. Implemented as a plugin in napari, VessQC leverages this powerful framework's capabilities for N-dimensional image visualization and interactive annotation. VessQC is designed to be model-agnostic, operating on datasets that include a raw image, a segmentation mask, and an uncertainty map.

To facilitate adoption, we provide open-source code for producing vessel segmentations with either pixel-wise or topology-aware uncertainty estimation (available at \\
github.com/SimPutt/VessQC-Supplementary). This separation keeps the plugin lightweight and independent of specific model architectures, while enabling users without prior segmentation or uncertainty maps to reproduce the full workflow end-to-end.

VessQC streamlines curation by prioritizing user attention to the most critical areas (see Figure \ref{fig:interface}). The workflow begins by ranking all volumes based on their highest uncertainty branch, ensuring users focusing on datasets with the most probable errors. Within a selected volume, the tool automatically identifies and highlights connected 3D regions that share similar uncertainty values, enabling direct navigation to these specific branches for review. From there, users can inspect the region as 3D rendered volume or slice by slice and compare the segmentation results against the raw image. Finally, manual corrections can be performed using various editing tools available in napari.

\begin{figure}[htb]
\begin{minipage}[b]{.48\linewidth}
  \centering
  \centerline{\includegraphics[width=4.0cm]{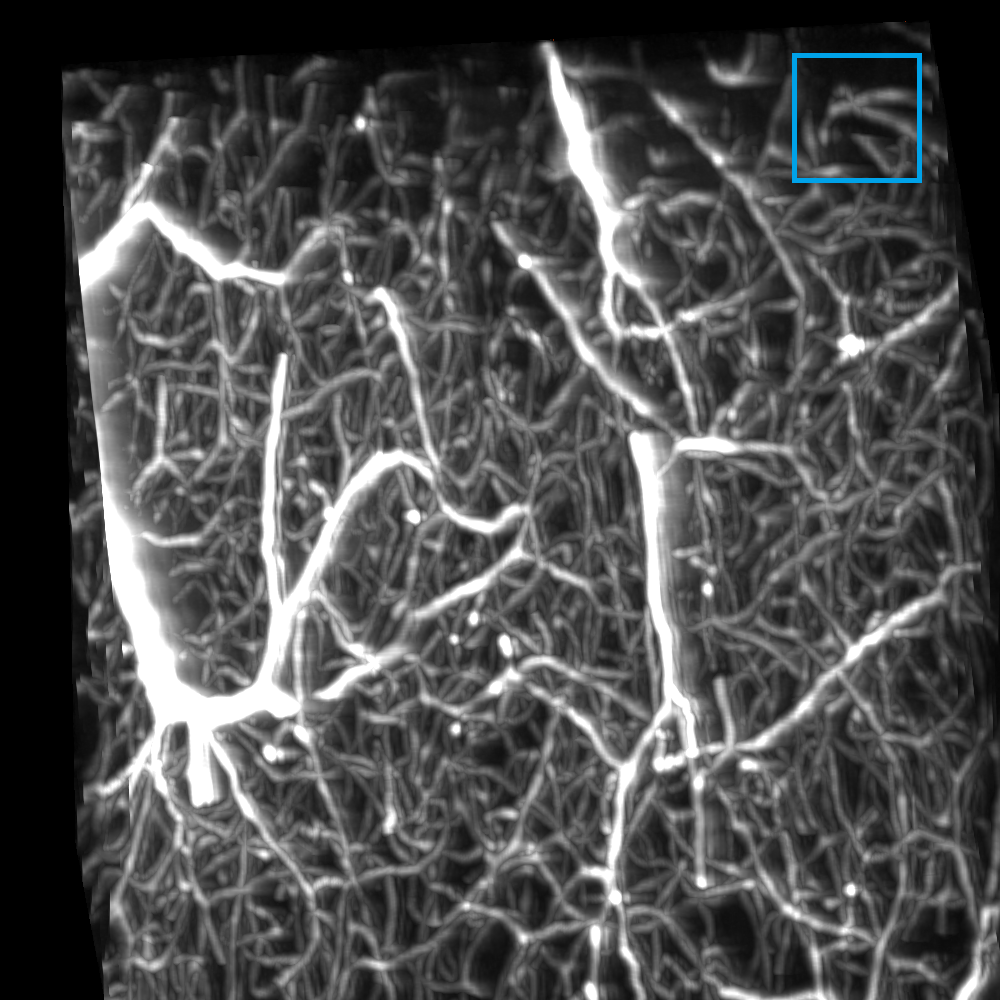}}
  \centerline{(a) 3D Raw-Data Overview}\medskip
\end{minipage}
\hfill
\begin{minipage}[b]{0.48\linewidth}
  \centering
  \centerline{\includegraphics[width=4.0cm]{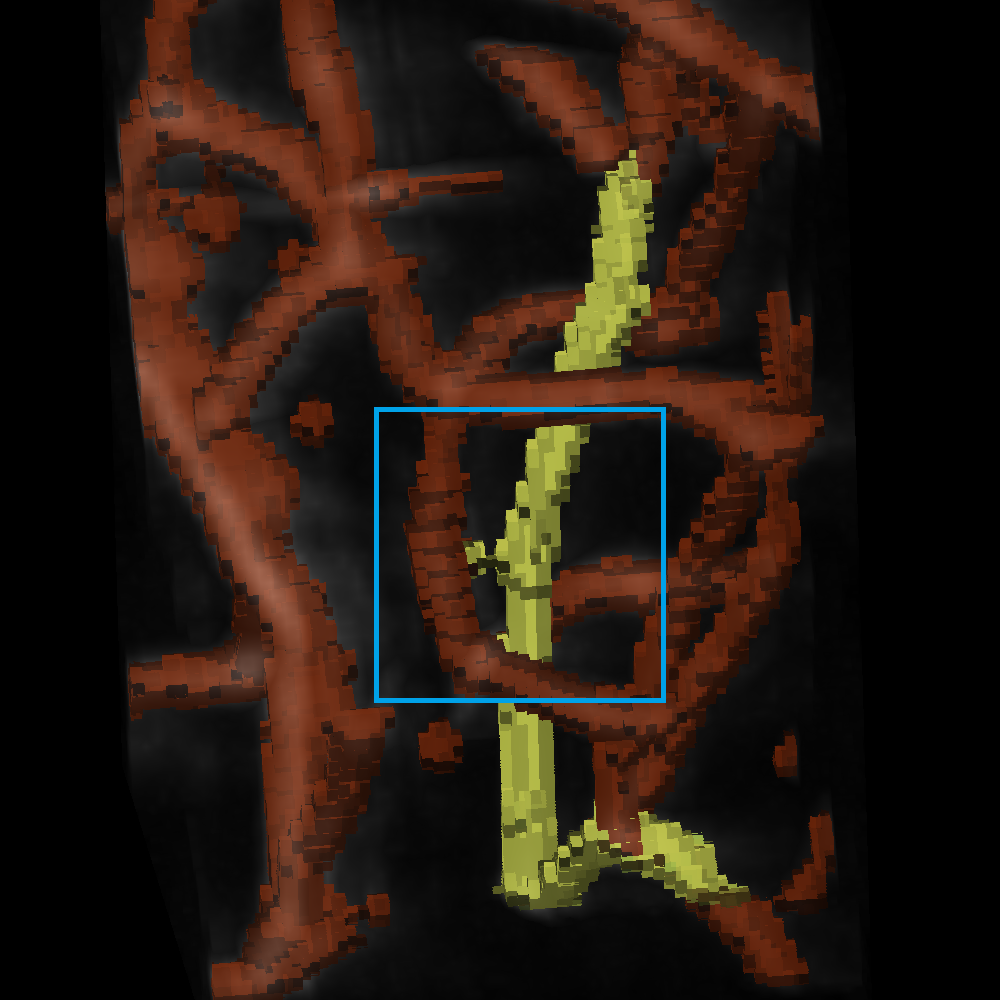}}
  \centerline{(b) Prediction (Pre-Curation)}\medskip
\end{minipage}
\begin{minipage}[b]{.48\linewidth}
  \centering
  \centerline{\includegraphics[width=4.0cm]{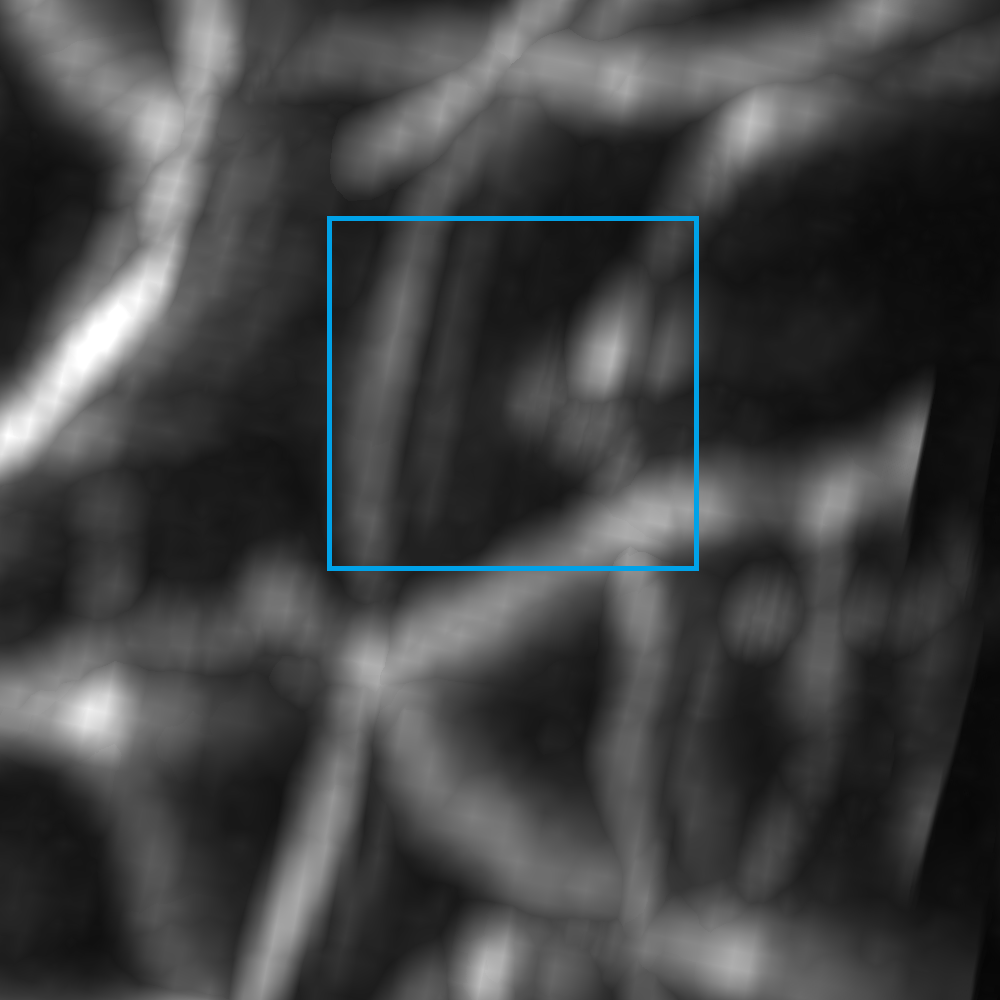}}
  \centerline{(c) Raw-Data Close-up}\medskip
\end{minipage}
\hfill
\begin{minipage}[b]{0.48\linewidth}
  \centering
  \centerline{\includegraphics[width=4.0cm]{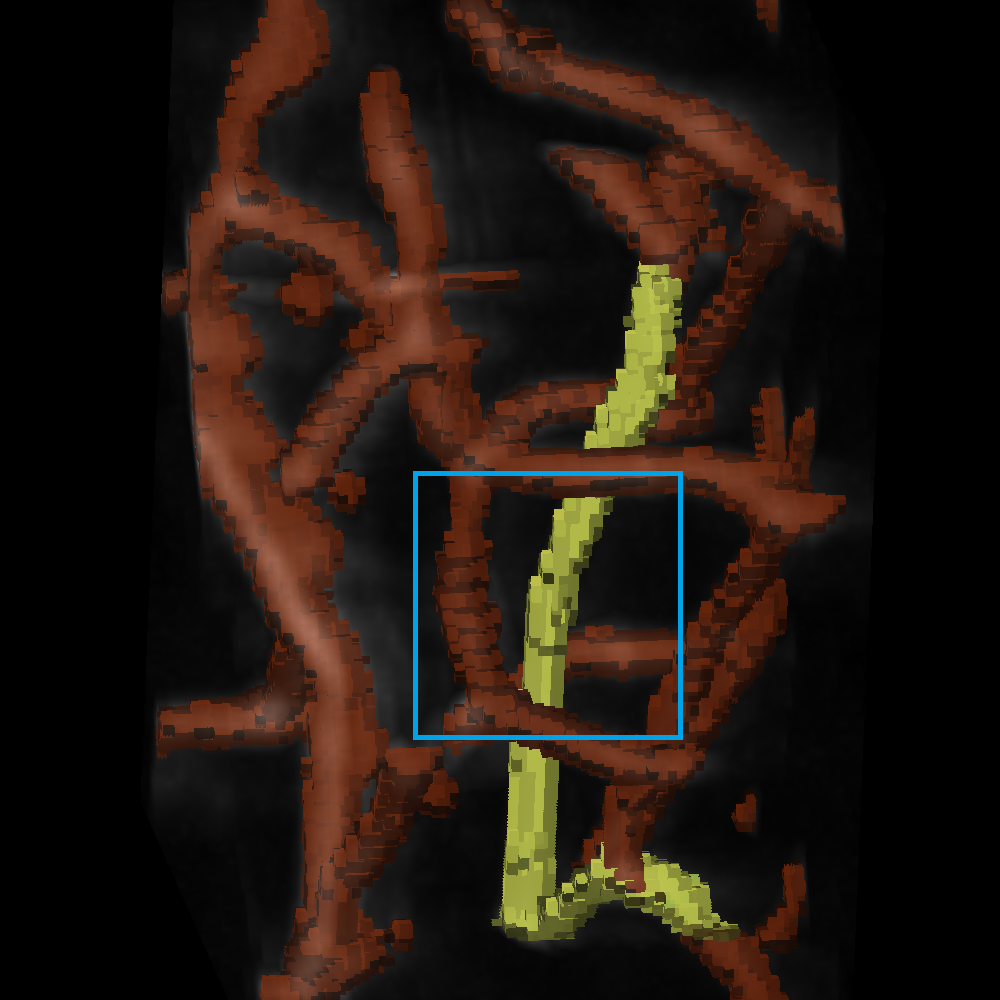}}
  \centerline{(d) Corrected Prediction}\medskip
\end{minipage}
\caption{Curation workflow in VessQC. A 3D overview of the vascular sample is shown (a), with the highlighted region containing a vessel flagged for high topology-based uncertainty ($\mathbin{\approx}0.8$). The pre-curation segmentation (red, in b) reveals a potential false merge between neighboring vessels, as indicated by the uncertainty cue (yellow). Inspection of the corresponding raw image data (c) confirms that the vessels are not physically connected. After manual correction within VessQC, the erroneous connection is removed and the two vessels are correctly separated (d).}
\label{fig:workflow}
\end{figure}

Figure \ref{fig:workflow} illustrates a representative curation workflow using a segmentation model with topology-aware uncertainty estimation. In this example, VessQC automatically highlights a vascular region with high topological uncertainty, indicating a potential false merge between two vessels. Guided by this uncertainty cue, the user inspects the corresponding raw data, confirms the segmentation error, and corrects it accordingly.


\section{Evaluation}
VessQC was evaluated on light-sheet microscopy images of 3D murine brain samples (CD31 labeled). The evaluation utilized paired 3D segmentation and topology uncertainty maps, which directly reflect potential connectivity inconsistencies in the predicted vascular network. Topology uncertainty was selected as it is highly relevant to structural error correction (implementation code is available at github.com/SimPutt/\hspace{0pt}VessQC-Supplementary/tree/main/\hspace{0pt}Topology-Aware-Uncertainty).

A preliminary user study was conducted to assess the efficiency and accuracy of VessQC. Participants corrected topological errors in segmented volumes of $100^3$ voxels, performing the task both with and without VessQC's uncertainty-guided workflow. Participants were instructed to locate and correct all topological inconsistencies, such as false merges or breaks in the vascular network, until they believed that all errors had been resolved. Both the total time per volume and the time required to detect and correct each individual error were recorded. The order of presentation was systematically varied across participants to mitigate potential learning or fatigue effects, and breaks were provided after every two tasks to maintain concentration.

Across all evaluated cases, curation assisted by VessQC led to a noticeably higher number of correctly identified and corrected topological errors compared to manual inspection alone.
The overall recall (accuracy of error detection) increased with statistical significance (two-tailed unpaired t-test, $\text{p}\mathbin{=}0.007$) from $0.670\,\pm\,0.173$ in the manual group to $0.940\,\pm\,0.098$ with VessQC guidance. This 1.40-fold improvement demonstrates that uncertainty-based guidance effectively directs user attention to ambiguous regions that are easily overlooked in conventional manual inspection. Furthermore, users maintained appropriate discernment, as no evidence of over-correction or false positive introduction was observed during the guided correction process.

On average, the entire correction process took $30 \pm 9 \text{ min}$ for VessQC and $22\pm 7 \text{ min}$ for the manual approach, making the VessQC process $35.2\%$ longer. However, this time difference was not statistically significant (two-tailed unpaired t-test, $\text{p}\mathbin{=}0.084$). Instead, the mean time required per correction performed was almost identical ($366\text{ s}$ for VessQC and $364\text{ s}$ for manual), confirming that the total time increase is attributable to the significantly higher error detection rate.

This observed time difference, despite the similar correction rate, can additionally be understood in the context of the cognitive demands of comprehensive manual review. In the absence of systematic guidance, such as in the manual setup, users face a significant challenge in maintaining sustained vigilance across complex datasets. This lack of an objective tracking mechanism, to verify which regions have been fully inspected, forces users to rely on satisficing heuristics (i.e., accepting a "good enough" solution rather than striving for completeness). Consequently, they are prone to a premature cessation of the inspection process, driven by a subjective sense of task completion rather than an objective verification of error clearance. This behavior explains the lower error detection rate and consequently, the faster completion time observed in the manual group, underscoring the performance limitations inherent in unguided data curation.

In larger-scale curation scenarios, where the complexity of the data volume and the difficulty of tracking necessary corrections are substantially amplified, the efficiency and quality advantages of VessQC are expected to become even more pronounced. In addition, user feedback collected during the study provided valuable insights into interface design and workflow ergonomics, which have already informed several refinements to the VessQC user interface aimed at streamlining interactions and accelerating manual curation tasks.

\section{Discussion \& Conclusion}
\label{sec:discussion}
In this work, we introduced VessQC, an open-source napari plugin designed to streamline the curation of 3D image segmentations in an efficient way. The tool directly addresses a critical bottleneck in real-life use cases in modern bioimage analysis: the time-consuming and often inefficient manual curation of segmentations in large volumetric datasets. 

VessQC is designed to be integrated into an iterative data-centric workflow (e.g., as proposed in \cite{cao2025rethinking}). This process typically begins with initial segmentation from a preliminary model (such as a pre-trained foundation model or a model trained on a small initial ground truth set) paired with a rough estimation of uncertainty. Users can then leverage VessQC to guide the curation of additional high-quality ground truth, which is subsequently used to fine-tune the model. The process can continue iteratively (model inference with uncertainty, curate additional high quality ground truth, further fine-tune the model) until satisfactory results are obtained. Essentially, VessQC translates the theoretical advantages of deep learning segmentation and uncertainty estimation into a tangible, practical tool for biomedical users, allowing them to systematically enhance the quality of their training data and resulting segmentations.

Our evaluation demonstrates that this approach is effective: VessQC significantly increases error detection accuracy ($\text{p}\mathbin{=}0.007$) without introducing a statistically significant overhead in total curation time ($\text{p}\mathbin{=}0.084$), validating the concept of efficiency through targeted guidance. Despite the limited scope of this initial study, the results suggest that uncertainty-based visual assistance can enhance both the sensitivity and reliability of human quality control in volumetric vessel segmentation. Future work will involve larger datasets, additional users, and more complex imaging conditions to quantitatively confirm these findings and better characterize the time-accuracy tradeoff, while also collecting user feedback to further optimize the tool.

While VessQC was developed and validated using 3D microscopy images of vascular networks, its core framework is structure-agnostic. Since it identifies connected regions of similar uncertainty and links them to their corresponding image areas, the same interface can be applied to a wide range of 3D datasets (i.e. cell segmentation).

Nevertheless, it is important to acknowledge that this uncertainty-guided approach is a powerful heuristic, not an infallible error detector. Not all segmentation mistakes will correspond to high uncertainty, and some high-uncertainty regions may be correctly segmented. VessQC should therefore be viewed as a tool to improve the efficiency of expert-led curation, where the final judgment remains with the user, rather than as a fully automated system.


In summary, VessQC addresses a critical need for efficient and accurate quality control in 3D bioimage analysis, representing the first open-source tool to integrate uncertainty estimation with interactive 3D curation for microscopy image segmentation. Our results demonstrate that uncertainty-guided visualization substantially enhances users’ ability to identify and correct topological segmentation errors without compromising efficiency. By prioritizing attention to uncertain and structurally critical regions, VessQC transforms manual quality control into a focused, data-centric process. Beyond vascular analysis, its structure-agnostic design makes it broadly applicable to diverse 3D bioimaging tasks. VessQC is released as open-source software (available at GitHub (github.com/MMV-Lab/VessQC), PyPI (pypi.org/project/VessQC/) and napari-hub (napari-hub.org/plugins/vessqc.html).

\section{Compliance with ethical standards}
\label{sec:ethics}
This study did not involve any new experiments on human or animal subjects. Therefore, ethical approval was not required.

\section{Acknowledgments}
\label{sec:acknowledgments}

S.P. is supported by the Federal Ministry for Research, Technology and Aeronautics (BMFTR) under reference number 01GQ2405A.
J.C., J.J.S.C., and L.K. are supported by the Federal Ministry for Research, Technology and Aeronautics (BMFTR) under reference number 161L0272.
The work of ISAS was additionally supported by the Ministerium für Kultur und Wissenschaft des Landes Nordrhein-Westfalen and Der Regierende Bürgermeister von Berlin, Senatskanzlei Wissenschaft und Forschung.
N.H. and D.M.H. are supported by the NIH–BMFTR Collaborative Research Project TopoVess (01GQ2405B, to J.C. and D.M.H.).
S.G. and C.C. acknowledge partial support from NSF CCF-2144901 and NIH R01NS143143.

\bibliographystyle{IEEEbib}
\bibliography{refs}

\end{document}